\documentclass{article}

\usepackage[preprint,nonatbib]{nips_2018}

\usepackage[utf8]{inputenc} % allow utf-8 input
\usepackage[T1]{fontenc}    % use 8-bit T1 fonts
\usepackage{hyperref}       % hyperlinks
\usepackage{url}            % simple URL typesetting
\usepackage{booktabs}       % professional-quality tables
\usepackage{amsfonts}       % blackboard math symbols
\usepackage{nicefrac}       % compact symbols for 1/2, etc.
\usepackage{microtype}      % microtypography
\usepackage{arydshln}

\usepackage{times}
\usepackage{graphicx}
\usepackage[square,numbers]{natbib}
\usepackage{algorithm}
\usepackage{algorithmic}
\usepackage{amsmath,amsfonts, amsthm}
\usepackage{mathrsfs}
\usepackage{bbm}
\usepackage{color}\usepackage{enumerate}
\usepackage{textpos}
\usepackage{xcolor}
\usepackage{xspace}
\usepackage{array}
\usepackage{url}
\usepackage{bm}

\usepackage{amsmath,amsfonts}
\usepackage{mathrsfs}
\usepackage{bbm}
\usepackage{color}\usepackage{enumerate}
\usepackage{textpos}
\usepackage{xcolor}
\usepackage{xspace}
\usepackage{array}
\usepackage{url}

\usepackage{booktabs}
\usepackage{subcaption}
\usepackage{lipsum}

\usepackage{sistyle}
\SIthousandsep{\,}
\newcommand{\BEAS}{\begin{eqnarray*}}
\newcommand{\EEAS}{\end{eqnarray*}}

\hypersetup{colorlinks = false, pdfborder = {0 0 0}}

\renewcommand\footnotemark{}

\title{Interpretable Set Functions}
\author{Andrew Cotter, Maya Gupta, Heinrich Jiang,  \\
 {\bf James Muller, Taman Narayan, Serena Wang, Tao Zhu}\thanks{\{acotter,mayagupta,heinrichj,muller,tamann,serenawang,tzhu\}@google.com} \\
  Google Research\\  
  1600 Amphitheatre Parkway, Mountain View, CA 94043\\
 \\
}

\begin{document}

\maketitle
\vspace{-2em}
\begin{abstract}
We propose learning flexible but interpretable functions that aggregate a variable-length set of permutation-invariant feature vectors to predict a label. We use a deep lattice network model so we can architect the model structure to enhance interpretability, and add monotonicity constraints between inputs-and-outputs. We then use the proposed set function to automate the engineering of dense, interpretable features from sparse categorical features, which we call semantic feature engine.  Experiments on real-world data show the achieved accuracy is similar to deep sets or deep neural networks, and is easier to debug and understand.   
\end{abstract}

\section{Introduction}
\label{intro}

We consider the problem of learning a function that acts on a variable-length set of un-ordered feature vectors. For example, for one of the experiments we will predict the sales $y \in \mathbb{R}$ of a product $x$ based on its $M(x)$ customer reviews, where for each customer review, there are $D$ features, such as that review's star rating and word count.  Recently, \citet{Smola:2017} showed that for such permutation-invariant countable sets $x$, all valid functions can be expressed as a transform of the average of per-token transforms: 
\begin{equation}
f(x) = \rho \left(\frac{1}{M(x)} \sum_{m=1}^{M(x)}  \phi({x_m})  \right) \label{eqn:smola}
\end{equation}
where $x_m \in \mathbb{R}^D$ is the $m$th token out of $M(x)$ tokens in the example set $x$, $\phi:\mathbb{R}^D \rightarrow \mathbb{R}^K$, and $\rho: \mathbb{R}^K \rightarrow \mathbb{R}$. (Note, we have changed their expression from a sum to an average to make the generalization of classic aggregation functions like the $\ell_p$ norm clearer; the two forms are equivalent because one of the $D$ features for each token $x_m$ can be the number of tokens $M(x)$). 

%For example, consider the simple approach of first averaging each of the $D$ features over the $M(x)$ tokens to produce an averaged $D$ features for the example $x$, which can then be treated as a standard machine learning problem with $D$ features. This can be expressed as (\ref{eqn:smola}) by adding the number of tokens $M(x)$ to the feature set, then setting $\phi(x_m) = \frac{1}{M(x)} x_m$, and then training $\rho$ on the $D$ averaged-features. 
%Of course, we generally expect to do better if we learn the $\phi(\cdot)$. 

\citet{Smola:2017} propose training neural networks representing $\phi$ and $\rho$, which can be jointly optimized, and they call \emph{deep sets}. This strategy inherits the arbitrary expressability of DNN's~\citep{Smola:2017}. Another formulation of (\ref{eqn:smola}) comes from \emph{support distribution machines}~\citep{Muandet:2012,Poczos:2012}, which in this context can be expressed as:
\begin{equation}
f(x) = \sum_{i=1}^N \alpha_i y^i \left(\frac{1}{M(x)} \sum_{m=1}^{M(x)} \left(\frac{1}{M(x^i)} \sum_{m'=1}^{M(x^i)} k(x_m, x^i_{m'}) \right) \right),
\end{equation}
where $\{(x^i, y^i)\}$ are training examples, $k(\cdot, \cdot)$ is a kernel, and those training examples with non-zero $\alpha_i$ play the role of finitely-sampled \emph{support distributions}.  Other work has also defined kernels for distributions derived from sets of inputs~\citep{Jebara:03}  (in less related work, some machine-learning algorithms have also been proposed for permutation-invariant inputs, without handling variable-length inputs, e.g.~\citep{Jebara:06}). Another related approach is that of \citet{hartford2016deep}, who create a deep neural network architecture which takes variable-sized bimatrix game as input and allows permutation invariance across the actions (i.e. rows and columns of the payoff matrix). 

\section{Interpretable Set Functions with Lattice Models}\label{proposed}
In this paper, we propose using the deep lattice network (DLN) function class~\citep{You:2017} for the $\phi$ and $\rho$ transforms in (\ref{eqn:smola}), which enables engineering more interpretable models than DNNs. This produces a new kind of DLN that we refer to as a \emph{DLN aggregation function}, which we abbreviate in some places as \emph{DLN agg function}. DLNs improves interpretability in two key ways: (i) the visualizability of the first layer of 1d calibrator curves, (ii) the ability to capture prior knowledge about global trends (aka monotonicity), as detailed in the following subsections. We also explain how bottle-necking (\ref{eqn:smola}) by setting $K=1$ improves debuggability.

\subsection{Calibrator Curves Promote Visual Understanding}
\begin{figure}[t]
\centering
\begin{tabular}{cc}
\includegraphics[width=0.3\textwidth,keepaspectratio=true]{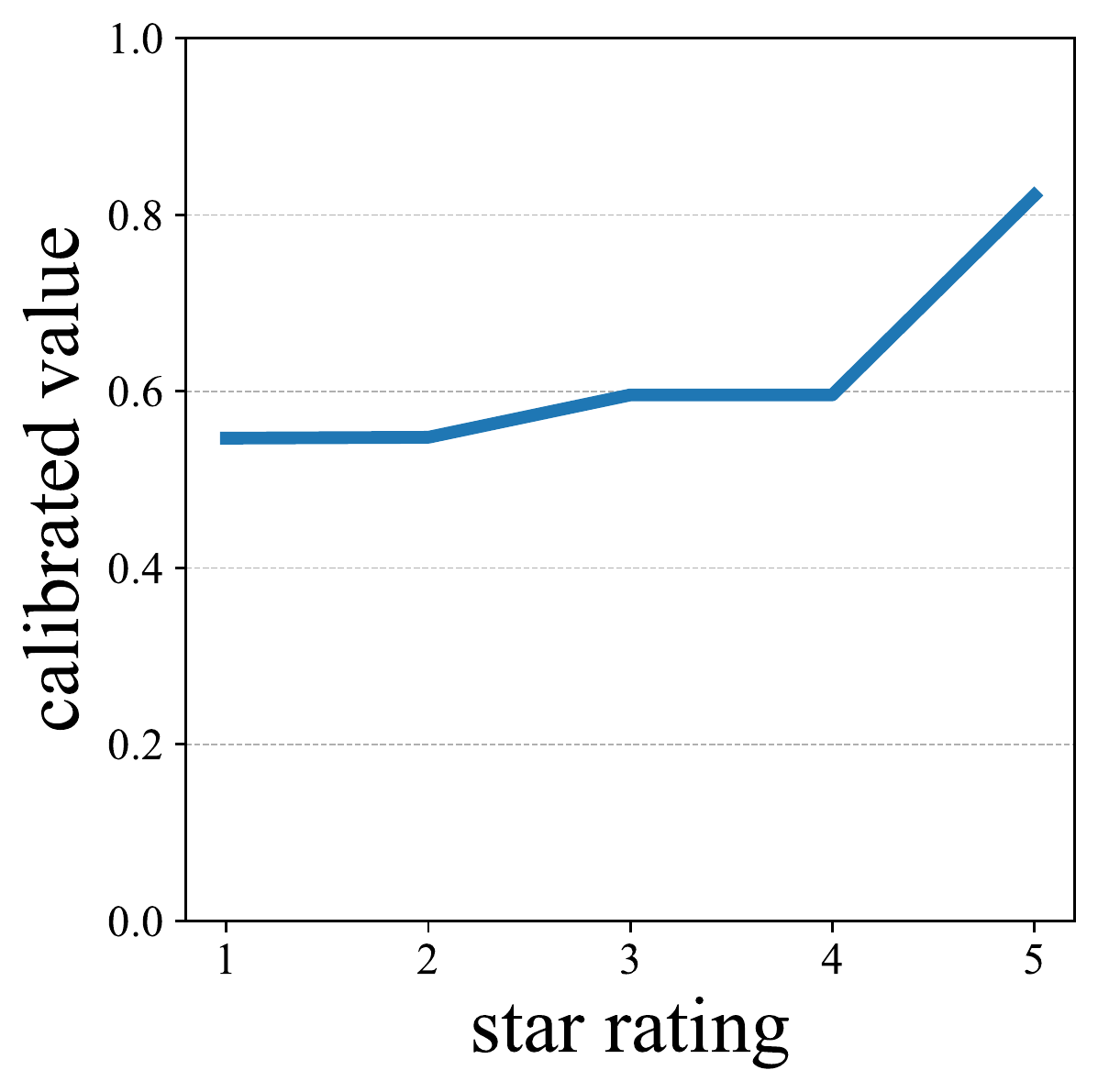} &
\includegraphics[width=0.3\textwidth,keepaspectratio=true]{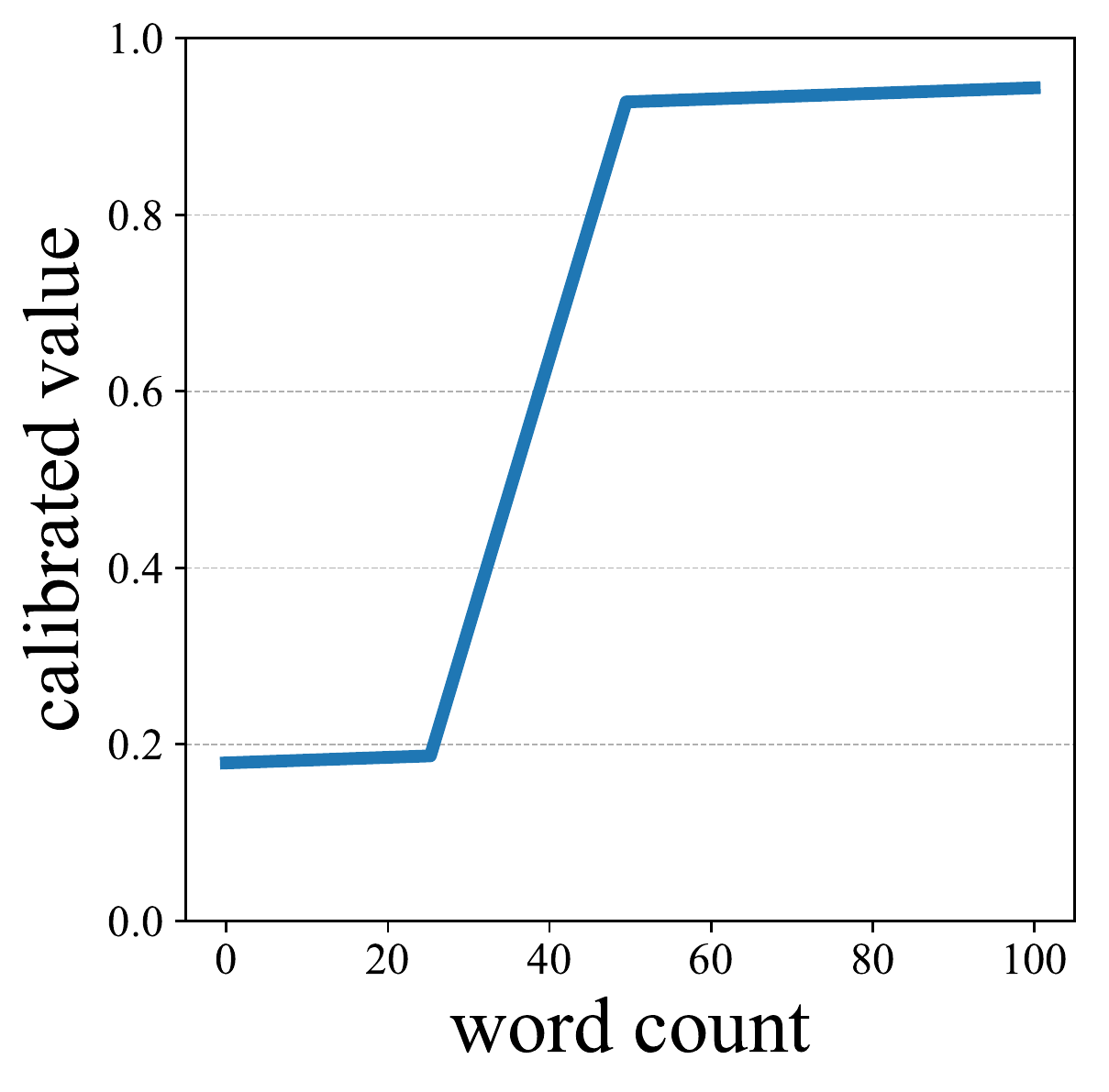}\\
\end{tabular}
\caption{Learned calibrator curves in the agg function for the dataset of Section \ref{experimentsPuzzle}. \textbf{Left:} One sees that the agg function has learned to treat 2 star reviews as just as bad as 1 star reviews, and similarly considers 3 and 4 star reviews of equal importance, but its main distinction is that anything lower than 5 stars is a bad sign. \textbf{Right:} This calibrator curve shows that agg function learned to treat reviews under 25 words as indistinguishably \emph{short}, and reviews over 50 words as equally usefully \emph{long}, and is linearly sensitive to reviews from 25-50 words long (around 1/4 of all reviews).}
\label{fig:calibrators} 
\end{figure}

The first layer of a DLN is a calibration layer that automates feature pre-processing by learning a 1-d nonlinear transform for each of the $D$ features using 1-d piecewise linear functions. The resulting 1-d calibrators are easy to visualize at and interpret (see Fig. \ref{fig:calibrators}). Specifically, we define the $k$th output of $\phi$ in (\ref{eqn:smola}) to take the form: 
\begin{equation}
\phi(x; \beta, \theta)[k] =  g_{\theta}\left(c_{\beta_1}(x_m[1]), \ldots, c_{\beta_D}(x_m[D])) \right), \label{eqn:calibrators}
\end{equation}
where each calibrator $c_{\beta_d}:\mathbb{R}\rightarrow [0,1]$ is an one-dimensional piecewise linear function, stored as a look-up table parameterized by vector $\beta_d$, and after each feature is calibrated, the $D$ features are fused together by other DLN layers represented here as $g_{\theta}:\mathbb{R}^{D} \rightarrow \mathbb{R}$ with parameters $\theta$. 

Such discriminatively-trained  per-feature transforms have been shown to be an efficient way to capture nonlinearities in each feature (e.g.~\citep{BalaBook,Jebara:2007,GuptaEtAl:2016}), and can also be framed as having a first layer to the model that is a generalized additve model (GAM)~\citep{HastieTibshirani:90}. Each piecewise-linear calibrator can equivalently be expressed as a sum of weighted, shifted ReLu's~\citep{You:2017}, but the look-up table parameterization enables monotonicity regularization. 

\subsection{Monotonicity Regularization Promotes End-to-End Model Understanding}
For many applications, there is domain knowledge that some features should have a monotonic impact on the output. Thus a particularly interpretable way to regularize is to constrain a model to capture such domain knowledge (see e.g.~\citep{Groeneboom:2014,Barlow:72,Jebara:2007,Daniels:2010,Sill:97,kotlowski2009rule,You:2017,Canini:2016,GuptaEtAl:2016}).  For example, when training a model to predict sales of a product $x$ given its  customer reviews $\{x_m\}$, we will constrain that if the star rating for the $m$th review is increased, the predicted product sales $f(x)$ should never go down.

Monotonicity constraints especially improve interpretability and debuggability for \emph{nonlinear} models, because no matter how complex the learned model is, the user knows the model respects their specified global properties, for example, that better reviews will never hurt predicted product sales. Monotonicity constraints are in general a handy regularizer  because the per-feature constraints can be set a priori by domain experts without needing to tune some how much regularization, and the resulting regularization is robust to domain shift between the train and test distributions. 

DLNs are a state-of-the-art function class for efficiently enabling monotonicity constraints~\citep{You:2017}. A DLN can alternate three kinds of layers: (i) calibration layers of one-dimensional piecewise linear transforms as in (\ref{eqn:calibrators}), (ii) linear embedding layers, and (iii) layers of multi-dimensional lattices (interpolated look-up tables) which enable nonlinear mixing of inputs. All three types of layers can be constrained for monotonicity, resulting  in end-to-end monotonicity guarantees (by composition). 

\subsection{Special Case: $K=1$ For Better Debuggability and Memory Usage}
We will show with the proposed DLN agg functions that for real problems we may be able to use a restricted architecture with just $K=1$ output from the $\phi$ function in (\ref{eqn:smola}), which has debuggability and memory advantages.

\textbf{Debuggability:} We have found that restricting $K=1$ such that $\phi:\mathbb{R}^D \rightarrow \mathbb{R}$  and $\rho: \mathbb{R} \rightarrow \mathbb{R}$ greatly aids interpretability, because the agg function becomes a visualizable 1-d transform $\rho$ after an average of $M(x)$ token values, and each token value $\phi(x_m)$ can be viewed and individually debugged, especially since each $\phi(x_m)$ is a smooth monotonic function and can be easily debugged with partial dependence plots.  In the customer reviews example, this makes it easy to quickly identify if a particular review is dominating the prediction, and if so, what it is about that review's $D$ features that is important. Limiting to $K=1$ still enables  learning variations of most of the classic aggregation functions, such as min, max, unnormalized weighted mean, geometric mean (which are also all monotonic with respect to the main feature). However, to express a normalized weighted mean of the form $(\sum_m x_m w_m)/(\sum_m w_m)$, requires the per-token $\phi$ function to produce two output values ($K=2$), one for the numerator and one for the denominator. 

\textbf{Memory Usage:} Using $K = 1$ or even just $K < D$  may make it possible to substantially reduce run-time memory, if there are a finite number of possible tokens $x_m$, because one can compute the $\phi(\cdot)$ offline for every possible token $x_m$, and then only store its $K$ values for each token. At runtime, one sees the exact tokens $\{x_m\}$ that are needed, retrieves the pre-computed  $\{\phi(x_m)\}$ values, and takes the average and applies $\rho$.

%ToDo(anyone): we might want to add some more classic aggregation functions to this list that I didn't yet verify, e.g. here are some more from Torra et al.~\citep{Torra:2010}:
%\begin{itemize} 
%\item harmonic mean (someone verify)
%\item t-norm (someone verify)
%\item Sugeno integral (someone verify)
%\item Choquet integral (someone verify)
%\item connection to Shapley value (someone investigate and leave %a comment if not interesting)
%\item any other standard aggregation functions we generalize?
%\end{itemize}

\section{Experiments on Sets of Feature Vectors}
\begin{figure}[t]
\centering
\begin{tabular}{cc}
\includegraphics[width=0.95\textwidth]{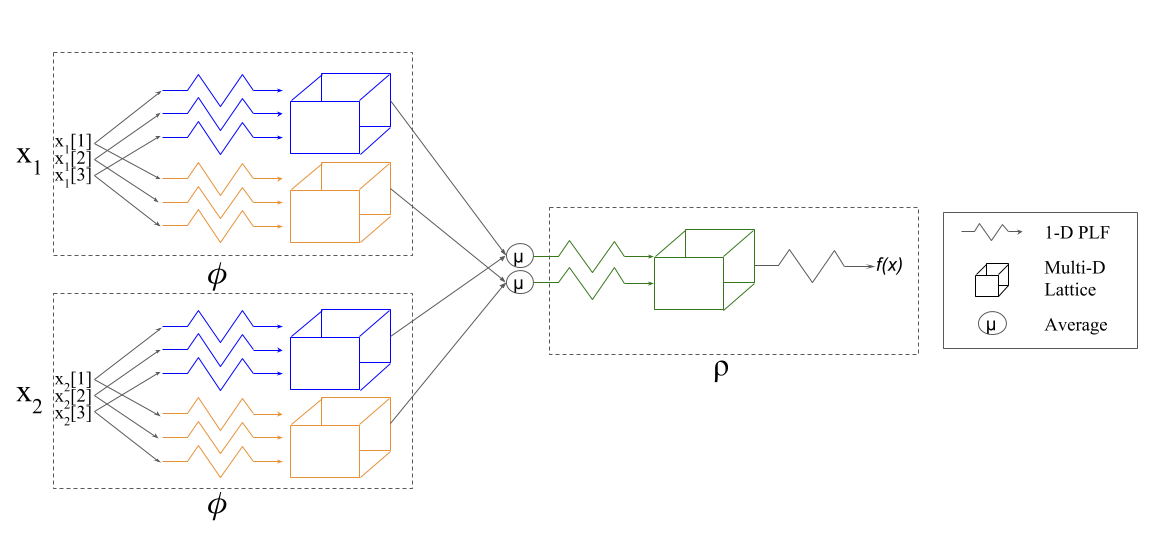}\\
\end{tabular}
\caption{Example set function architecture with $M(x)=2$ tokens, $D=3$ features per token, and $K=2$ intermediate dimensions. Note that the same function $\phi: \mathbb{R}^3 \rightarrow \mathbb{R}^2$ is applied to each input token $x_m$, before the outputs are averaged along each of the $K=2$ dimensions and fed into $\rho: \mathbb{R}^2 \rightarrow \mathbb{R}$ to produce the final output.}
\label{fig:architecture} 
\end{figure}

We demonstrate the proposed aggregation functions with two real-world case studies (more experiments in Section \ref{sec:experimentsSFE}). The DLN Agg function architecture we use is illustrated in Fig. \ref{fig:architecture}. 

\subsection{Implementation Details}
For all the experiments in this paper, as show in Fig. \ref{fig:architecture}, we use a 6-layer DLN Agg function archiecture composed of  $K$ calibrated lattice models for $\phi$, followed by an average, and then $\rho$ is composed of a $K$-dimensional calibrated lattice model followed by a final one-dimensional calibration layer. All DLN layers are differentiable, and thus we jointly train the parameters of $\rho$ and $\phi$ using backpropagation of gradients as in (\ref{eqn:smola}). See Appendix A in the supplemental for more details on our implementation, initialization, and optimization of $\rho$ and $\phi$, which largely follow the descriptions in other recent papers on lattice models~\citep{You:2017,Canini:2016,GuptaEtAl:2016}. 

We will provide open-source Tensor Flow code to implement the proposed agg functions, building on the DLN layers and monotonicity projection operators of the open source Tensor Flow Lattice package (github.com/tensorflow/lattice).

We compare to \emph{deep sets}~\citep{Smola:2017}: for all the all deep sets comparisons, we model $\phi$ and $\rho$ each as 3-hidden-layer fully-connected DNNs implemented in TensorFlow, with the number of hidden nodes for each layer fixed to be the same, and $K$ was also set to that value. We then trained using the ADAM optimizer \cite{kingma2014adam}. The hyperparameters validated over were learning rate, the number of units in each hidden layer, and number of training iterations.
%See Appendix C for more details about implementation details and hyperparameters that were validated. 

\subsection{Case Study: How Customer Reviews Affect Product Sales}\label{experimentsPuzzle}
This case study illustrates the interpretability of the proposed DLN aggregation. The goal is to understand how different aspects of product reviews affect product sales. The data, from a luxury goods company [name redacted for blind review], will be made publicly available on Kaggle. The training set label is number of sales of the product $x$ over a six month window, and the training set are all the products in stock during that time period, and the training features are derived from all product reviews posted at the end of that time period. The validation and test sets are analogous, but for the next two six month periods, respectively. This produces $N=156/169/200$ train/validation/test samples, which are non-IID due to the time shift and because $156$ of the validation and test products are the same as the training products, albeit with the statistics collected over different time periods (the rest are newly-released products). Each product $x$ is described by $M(x) \in [1-55]$ customer reviews, and the $D=3$ features (i) the star rating of each review, (ii) the word count for the review, and (iii) the number of reviews that product got $M(x)$.  While tiny, this real-world example is excellent for analyzing and comparing flexibility-regularization trade-offs. For the proposed aggregation functions, we constrained the predicted sales to be monotonically increasing in the star rating, and in the number of reviews $M(x)$ (which signifies popularity).

Results are given in Table \ref{tab:reviews}. DLN agg functions are able to achieve the best performance on the test set in this example; deep sets, along with the naive linear regression baseline, perform substantially worse. The large gap in performance is likely due to two factors: (1) the small size of the dataset and (2) the test set being non-IID with respect to the training set. Both of these create advantages for simpler and more regularized DLN models.

%Notably, the linear function learned on the averaged features (top row) actually learns a negative coefficient around $-17$ on star rating. This appears to be because the strongest positive indicator for sales is actually the number of reviews $M(x)$, but the larger the number of reviews, the more likely there will be a bad review or two that brings down the average star rating.  

\begin{table*}[t]
\caption{Mean Absolute Error Estimating Product Sales From Reviews}
\label{tab:reviews}
\begin{center}
\begin{tabular}{lllllll}
\toprule
 & Train Set & Validation Set & Test Set \\
\midrule
Averaged Aggregation, Linear Fusion & 3308 & 3771 & 8221 \\
Deep Sets & 561.8 & 2377 & 8323 \\
DLN Agg, $K=1$ & 3054 & 3454 & 7502 \\
DLN Agg, $K>1$ & 2646 & 2894 & 7737 \\
\bottomrule
\end{tabular}
\end{center}
\end{table*}

\subsection{Case Study: Predicting User Intent}
For this binary classification problem from a large internet services company, the goal is to predict if a given query $x$ (a string containing multiple words) is seeking a specific type of result. We use $N = 500,000$ examples, which we split randomly into training/validation/test sets in 8/1/1 proportions. Each query $x_i$ is broken into $M(x) \in \left\{1,2,\dots,70\right\}$ ngrams, and each ngram has $D=10$ corresponding pieces of information. Two of the $D=10$ features should be positive signals for the intent (e.g. what percentage of users who issued that ngram in the past were seeking this result type), and their effect is constrained to be monotonic. The other eight features are conditional features, e.g. how popular the ngram is, or the order of (number of terms in) the ngram. 

Table \ref{tab:intent} shows that specific set function models are significantly better than models on pre-averaged features. Furthermore, DLNs with $K>1$ are the best performing of the set function models. The $K=1$ DLN performs very similarly to the Deep Sets approach.

%For example, the query [taj mahal] could be seeking information about the Indian monument or the blues singer, similarly [coffee] is likely seeking to learn about nearby coffee shops, but [is coffee carcinogenic] is not. 

\begin{table*}[t]
\caption{Accuracy for Classifying User Intent}
\label{tab:intent}
\begin{center}
\begin{tabular}{lllllll}
\toprule
 & Train Set & Validation Set & Test Set \\
\midrule
Averaged Aggregation, Linear Fusion & 0.623 & 0.609 & 0.610 \\
Averaged Aggregation, DNN Fusion & 0.634 & 0.623 & 0.624 \\
Deep Sets & 0.662 & 0.644 & 0.643 \\
DLN Agg, $K=1$ & 0.653 & 0.644 & 0.643 \\
DLN Agg, $K>1$ & 0.674 & 0.648 & 0.646 \\
\bottomrule
\end{tabular}
\end{center}
\end{table*}

\section{Semantic Feature Engine}
We propose applying set function learning to handle sparse categorical variables in a debuggable and stable way, an approach we call the \emph{Semantic Feature Engine} (SFE).  A common approach to sparse categoricals is to create a Boolean feature for each possible category, and either use these directly as predictors or train an embedding. These strategies work well, but have poor interpretability, debuggability, and can be highly variable across retrainings causing unwanted churn and instability~\citep{Cormier:2016}.  By contrast, our proposed SFE converts the sparse categoricals into one dense, understandable feature that is an estimate of $E[Y]$, for some label $Y$. 

For example, suppose the goal is to produce a classifier that predicts if a movie $x$ will be rated PG (suitable for all-ages), and we want to use information about the movie's actors. SFE produces a feature that is an estimate $P\textrm{(rated PG | \{actors\}} \in x)$, which will have the semantic meaning of an \emph{actor prior} feature. This feature could then be combined with other dense and meaningful features, such as the movie's budget or studio's previous track record, to produce a final model that is both powerful and interpretable. 

In general, SFE produces a feature that is an estimate of some label $y \in \mathbb{R}$ given some set $x$. In simple cases one can simply form the point estimate $E[ Y | x ]$. More generally, $x$ may not have occurred often  enough in the training data to derive a straightforward point estimate; for example, many movies have sets of actors who have never appeared together before. To address this, the key idea of SFE is to convert $x$ into a set of tokens $\{x_m\}$, estimate $E[ Y | x_m ]$ for each token, then learn the best aggregation of the set of $M(x)$ estimates to form the best overall estimate $E[ Y | x ]$, which can be used as a feature in a bigger model. See Appendix B (8.2) in the Supplemental for a complete worked example.

% Maya says: I don't think we need any of this
%In addition to the training set of $\{x^i, y^i\}$ pairs used to the train the set function,
%The SFE also requires a \emph{token table training set} (which can be the same as the set used to learn the set function) that is used to compute the feature values for each unique token $\{x_m\}$. For example, to create the SFE feature $\textrm{P(rated PG | \{actors\}} \in x)$, we might compute $D=2$ token features for each actor, where $x_m[1]$ is the number of times the actor is in a PG movie, and $x_m[2]$ is the number of times the actor is in any movie.

%For each SFE feature, one must define (see below for an example):

\textbf{Tokenization and Fallback Rules:} If $x$ is not a single element, one must choose a tokenization rule to produce a set of tokens $\{x_m\}$ from a given example $x$. For example, for text, a standard tokenization is to break the text into $q$-order ngrams up to some max order $Q$. For a set of categorical variable, such as \{actors\}, one can tokenize it into all $k$-tuples up to some max subset size $K$. If $x$ is a pair of sets (e.g. the actors in a candidate movie to recommend, \emph{and} the list of all actors in movies the user has previously watched), tokens can be crosses or set differences across the pair. In addition, we suggest adding fallback rules if the $D$ token values are missing for a given token. For sets of categorical variables, our fallback is to iteratively consider size $k-1$ subsets for any categories not contained in an existing token of subset size $k$ until each category appears in at least one token if possible (see Appendix B for full examples). The last fallback is always to set the $D$ token values to missing.  

\textbf{Training Sets:} The SFE needs an \emph{SFE token training set} of $\{x^i, y^i\}$ pairs  to train the per-token estimates of $E[Y | x_m]$. This can be the same training set as used to learn the SFE aggregation function, and in fact this simple approach often works well. Using different training sets can reduce overfitting. One may also prefer to use different labels, for example, training the SFE per-token estimates of $E[Y | x_m]$ on a large (but noisy) dataset of clicks, but training the aggregration function to produce $E[Y | x]$ on a smaller, cleaner, human-labeled dataset. 

\textbf{Token Table Building:} Given tokenization rules, build a \emph{token table} by iterating through the SFE token training set to populate a table with empirical estimates of $E[Y | x_m]$, and possibly other aggregate statistics such as how often the token was seen in the training data. One can also store non-aggregated token-specific information, such as a token's subset size, in a table during the pre-computation phase; alternatively, these values could be populated at training time. Note that when training the set function, one can also add example-specific details (e.g. for actors, their salary or number of lines in the given movie) to provide additional information on how to weight the different token values. Cumulatively, these techniques provide the $D$ features per token $x_m$.

\textbf{Token Table Filtering:} To reduce table size and improve the statistical significance of the SFE signal, one should filter the token table before learning the SFE aggregation function. Two filtering rules that we have found useful are: (i) a count threshold (that is, if there are too few examples of a specific token in the token training set, then it should be dropped from the token table), (ii) a confidence interval threshold (that is, if one of the token features is an estimate of a target label $y$, and the confidence interval of that estimate is too big, then it should be dropped from the token table). Compared to count-thresholding, confidence interval filtering will keep more lower-frequency tokens that have high label-agreement amongst their occurrences. 

\textbf{Learn an Aggregation Function:} Given the tokens and their $D$ token values, train a set function $f(x)$ as per (\ref{eqn:smola}) to estimate the label $E[ Y | x ]$, with monotonicity  constraints on the token features based on domain knowledge. Generally, it makes sense to constrain the point estimate feature $E[ Y | x_m ]$ to have a monotonic impact on the aggregation function's prediction of $E[ Y | x ]$.  

\textbf{Extended Example:} Let $y = 1$ if the movie is PG and $0$ otherwise. Let $x$ represent the set of actors, and suppose a new  movie comes out with actors \{Alice, Bob, Carol and David\}. Suppose the tokenization rule is to use all subsets of up to size 3, with a fallback to smaller subsets if their parent sets are not in the SFE token table. This rule produces tokens \{ \{Alice, Carol\}, \{Bob, Carol\}, \{David\}\}, meaning that no set of 3 actors appeared in the table, nor did David appear with any of the other actors in the table. Suppose our $D=2$ token features are (i) the percentage of movies with that set of actors that are PG, and (ii) the number of movies with that set of actors. Then we might find that $x_1[1] = 0.62, x_1[2] = 67$, $x_2[1] = 0.70, x_2[2] = 51$, and $x_2[1] = 1, x_2[2] = 1$. We can then apply a set function $f(x)$ that was trained on all pre-2017 movies and was constrained to be monotonically increasing in the first feature, which might produce a \emph{movie title prior} probability of $f(x) = 0.68$, which can then be combined with other features for a final prediction of the movie's rating.

\textbf{Separability:} Here, we have separated the overall model into three parts: (i) the token table that stores token values, (ii) the set function that combines the $D$ token features across the $M(x)$ tokens for each example, (iii) the follow-on model which might take many SFE features as inputs. This separability has two key advantages in practice. First, each of these parts can has semantic meanings that aids interpretability and debuggability.  Second, each of these three parts can be refreshed or improved independently, which reduces churn~\citep{Cormier:2016} and system complexity. That said, jointly training the SFE set function and the follow-on classifier could result in additional metric gains.

\section{Experiments using the Semantic Feature Engine To Create Sets} \label{sec:experimentsSFE}
We also evaluate the performance of our set function learning approach when applied as part of the Semantic Feature Engine to sparse categorical predictors; that is, precomputing statistics about each category and learning a set function over token statistics. We show that as in the earlier experiments, it outperforms Deep Sets in learning the best set function over the tokens. We also compare it to a strategy of directly learning a deep neural network (DNN) on a multi-hot encoding of the categories and find that it tends to perform similarly well, with the added benefits of interpretability and debuggability.

For each experiment, our original predictors are a variable-length set of categories, and our SFE tokens are subsets of categories. In particular, we compute $D=6$ features for each token that is then fed into the set function: (a) the average label, computed over the training data, for a token; (b) how frequently the token appears; (c) the size of the subset $k$ that the token represents; (d) whether the token fully matches a set's list of categories; (e) the number of categories in the set; and (f) the number of tokens generated from the set. Note that these fall into three buckets: (a) is the direct estimate of the label's value for a token; (b), (c), and (d) are token-specific features that can provide information on how much to weight the tokens; and (e) and (f) are set-specific features (the same for all tokens in an example) to help calibrate the outputs and nature of the aggregation.

For the DNN comparison, we use standard TensorFlow embeddings, first creating a tf.feature\_column.embedding\_column over the raw categories (that appear in the train set), then feeding that into a tf.estimator.DNNClassifier with two hidden layers before a final softmax layer. We optimize the ADAM learning rate, number of epochs, and size of the embedding and hidden layers over the validation set. New validation-set or test-set categories are ignored; only embeddings from categories that also appear in the training set are used in evaluation.

\begin{table*}[t]
\caption{Accuracy for Classifying Facial Attractiveness}
\label{tab:celeba}
\begin{center}
\begin{tabular}{lllllll}
\toprule
 & Train Set & Validation Set & Test Set \\
\midrule
DNN on Attributes & 0.793 & 0.789 & 0.790 \\
Deep Sets & 0.794 & 0.786 & 0.785 \\
DLN Agg $K=1$ & 0.795 & 0.785 & 0.785 \\
DLN Agg $K>1$ & 0.795 & 0.786 & 0.786 \\
\bottomrule
\end{tabular}
\end{center}
\end{table*}

\subsection{CelebA}
There are $N = 202,599$ images of faces~\cite{liu2015faces}, which we randomly split 70/10/20 into a train/validation/test set. Each face is described by 40 binary attributes, such as whether the subject has blond hair, earrings, or a mustache. There is also a Boolean feature for whether the face was judged to be attractive. We treat the problem as a binary classifier of predicting whether a face is labeled as attractive based on its attributes, and use the Semantic Feature Engine to generate conditional probabilities of attractiveness for all subsets of attributes, whose estimated values have confidence intervals of 0.2 or under.

Results in Table \ref{tab:celeba} show that DLNs on aggregated token subset information, including the very simple $K=1$ model, perform better than Deep Sets models with substantially higher dimensional $\phi$ functions. DNNs directly on the attributes, however, perform the best of all models considered, with nearly 0.4\% higher test accuracy than the best-performing DLN.

\begin{table*}[t]
\caption{Accuracy and Ranking Precision for Classifying Recipe Cuisine}
\label{tab:recipes}
\begin{center}
\begin{tabular}{lllllll}
\toprule
 & Train Acc & Validation Acc & Test Acc & Test Prec@1 & Test Prec@3 \\
\midrule
DNN on Attributes & 0.986 & 0.974 & 0.974 & 0.728 & 0.883 \\
Deep Sets & 0.981 & 0.973 & 0.973 & 0.707 & 0.888 \\
DLN Agg $K=1$ & 0.991 & 0.974 & 0.973 & 0.710 & 0.880 \\
DLN Agg $K>1$ & 0.984 & 0.974 & 0.974 & 0.729 & 0.890 \\
\bottomrule
\end{tabular}
\end{center}
\end{table*}

\subsection{Cuisine Classification from Recipe Ingredient List}
The recipes dataset (\url{www.kaggle.com/kaggle/recipe-ingredients-dataset}) consists of $N = 39,774$ recipes represented by their list of ingredients, and the cuisine they come from. It is randomly split 70/10/20 into a train/validation/test set. We build a model that takes a list of ingredients and a cuisine and acts as a binary classifier that estimates if it's a correct match, that is, the model outputs an estimate of $P\left\{\textrm{cuisine c is correct} \mid \textrm{ingredients list}, \textrm{cuisine c}\right\}$. There is a fixed set of 20 possible cuisines, so we create one positive and nineteen negative training samples from each row of the original dataset. Note that the problem can also be thought of as a multiclass classifier; for that reason, we also report the multiclass metrics precision@1 and precision@3 (which compute how often the correct cuisine's score was the top or among the top-3 predictions for all candidate cuisines for a given recipe).

We start by doing some basic pre-processing on the ingredients, such as converting to lower case and word-stemming, to make it more likely that the equivalent ingredients are identified; we will upload a Kaggle kernel with the details. Then, we use SFE to tokenize the resulting ingredient set with each item crossed with each cuisine, see Appendix B (8.2) for a complete example.  We filter out token values for any sets of ingredients that appear fewer than 5 times in the training data. We consider subsets of ingredients up to size 3; we cross-validated max subset sizes up to 5. Using all subsets is not feasible; many recipes have over 20 ingredients and the longest has 65, meaning that hundreds or even thousands of tokens are being aggregated for some examples even with a max subset size of 3. Another unique feature of the recipes dataset compared with the other benchmarks is that the vocabulary is very large. There are over $6,000$ unique ingredients, meaning that many ingredients in the test set do not appear or barely appear in the training set.

As Table \ref{tab:recipes} shows, DLN agg functions with $K>1$ (in particular, $K=4$ here) perform the best of all models on the precision metrics, while DNNs are slightly higher on binary classifier accuracy. The $K=1$ DLN performs similarly to the higher-complexity Deep Sets model.

\begin{table*}[t]
\caption{Mean Squared Error for Predicting Wine Quality}
\label{tab:wine}
\begin{center}
\begin{tabular}{lllllll}
\toprule
 & Train Set & Validation Set & Test Set \\
\midrule
DNN on Attributes & 7.04 & 7.39 & 7.20 \\
Deep Sets & 7.29 & 7.45 & 7.26 \\
DLN Agg $K=1$ & 7.04 & 7.37 & 7.19 \\
DLN Agg $K>1$ & 7.02 & 7.35 & 7.19 \\
\bottomrule
\end{tabular}
\end{center}
\end{table*}

\subsection{Wine}
The wine dataset (\url{www.kaggle.com/zynicide/wine-reviews}) consists of $N=80,100$ different wines, along with their quality, price, country of origin, and a set of descriptive terms culled from reviews (out of a total set of 39 possible adjectives such as complex, oak, and velvet). We focus on predicting the quality, which is scored on a 100-point scale, using the set of review adjectives. As discussed in the SFE section, this \emph{review prior} score could be used in a follow-on interpretable model that also incorporated the other features such as price, though we do not do so here. We consider all subsets and require sets of adjectives to appear at least 32 times in the training data before entering the token table.

Table ~\ref{tab:wine} shows that DLNs perform best on the wine dataset, with even the $K=1$ DLN performing better than the DNN. Deep Sets struggles here, performing far worse than either of the competitors.

\section{Conclusions}
We have shown that we can learn DLN aggregation functions over sets that provide similar accuracy to deep sets \cite{Smola:2017}, but provide greater interpretability due to three key aspects. First DLN agg functions enable monotonicity constraints, providing end-to-end high-level understanding and greater predictability of even highly nonlinear models. Second, the first-layer of 1-d per-feature calibrator functions can be visualized and interpreted.  Third, we showed that we can simplify the middle layer to a per-token score before averaging ($K=1$), with slight or no loss of accuracy on real-world problems. This greatly aids debuggability as it makes it easier to determine which tokens are most responsible for the output, and whether any of the per-token scores are noisy or suspicious. 

We show that learning on sets is broadly applicable with our \emph{semantic feature engine} proposal, which converts highly sparse features into a dense estimate of $E[Y|x]$.  Our experiments show these estimates were similar in accuracy to applying a DNN to the sparse features. The main advantage of the SFE over the DNN is its greater interpretability and debuggability. We expect SFE will also show greater stability and less churn if re-trained.  SFE features can be combined with other information in follow-on DLN models, with monotonicity regularization on the SFE features, which makes the follow-on DLN more interpretable and stable over re-trainings.

\clearpage
\newpage

\bibliographystyle{abbrvnat}
\bibliography{references}

\begin{thebibliography}{22}
\providecommand{\natexlab}[1]{#1}
\providecommand{\url}[1]{\texttt{#1}}
\expandafter\ifx\csname urlstyle\endcsname\relax
  \providecommand{\doi}[1]{doi: #1}\else
  \providecommand{\doi}{doi: \begingroup \urlstyle{rm}\Url}\fi

\bibitem[Barlow et~al.(1972)Barlow, Bartholomew, Bremner, and Brunk]{Barlow:72}
R.~E. Barlow, D.~J. Bartholomew, J.~M. Bremner, and H.~D. Brunk.
\newblock \emph{Statistical inference under order restrictions; the theory and
  application of isotonic regression}.
\newblock Wiley, New York, USA, 1972.

\bibitem[Canini et~al.(2016)Canini, Cotter, Fard, Gupta, and
  Pfeifer]{Canini:2016}
K.~Canini, A.~Cotter, M.~M. Fard, M.~R. Gupta, and J.~Pfeifer.
\newblock Fast and flexible monotonic functions with ensembles of lattices.
\newblock \emph{Advances in Neural Information Processing Systems {(NIPS)}},
  2016.

\bibitem[Cormier et~al.(2016)Cormier, {Milani Fard}, and Gupta]{Cormier:2016}
Q.~Cormier, M.~{Milani Fard}, and M.~R. Gupta.
\newblock Launch and iterate: Reducing prediction churn.
\newblock \emph{Advances in Neural Information Processing Systems {(NIPS)}},
  2016.

\bibitem[Cotter et~al.(2016)Cotter, Gupta, and Pfeifer]{cotter2016light}
A.~Cotter, M.~R. Gupta, and J.~Pfeifer.
\newblock A {Light Touch} for heavily constrained {SGD}.
\newblock In \emph{29th Annual Conference on Learning Theory}, pages 729--771,
  2016.

\bibitem[Daniels and Velikova(2010)]{Daniels:2010}
H.~Daniels and M.~Velikova.
\newblock Monotone and partially monotone neural networks.
\newblock \emph{{IEEE} Trans. Neural Networks}, 21\penalty0 (6):\penalty0
  906--917, 2010.

\bibitem[Duchi et~al.(2011)Duchi, Hazan, and Singer]{duchi:2011}
J.~Duchi, E.~Hazan, and Y.~Singer.
\newblock Adaptive subgradient methods for online learning and stochastic
  optimization.
\newblock \emph{Journal Machine Learning Research}, 12:\penalty0 2121--2159,
  2011.

\bibitem[Groeneboom and Jongbloed(2014)]{Groeneboom:2014}
P.~Groeneboom and G.~Jongbloed.
\newblock \emph{Nonparametric estimation under shape constraints}.
\newblock Cambridge Press, New York, USA, 2014.

\bibitem[Gupta et~al.(2016)Gupta, Cotter, Pfeifer, Voevodski, Canini, Mangylov,
  Moczydlowski, and Esbroeck]{GuptaEtAl:2016}
M.~R. Gupta, A.~Cotter, J.~Pfeifer, K.~Voevodski, K.~Canini, A.~Mangylov,
  W.~Moczydlowski, and A.~V. Esbroeck.
\newblock Monotonic calibrated interpolated look-up tables.
\newblock \emph{Journal of Machine Learning Research}, 17\penalty0
  (109):\penalty0 1--47, 2016.
\newblock URL \url{http://jmlr.org/papers/v17/15-243.html}.

\bibitem[Hartford et~al.(2016)Hartford, Wright, and
  Leyton-Brown]{hartford2016deep}
J.~S. Hartford, J.~R. Wright, and K.~Leyton-Brown.
\newblock Deep learning for predicting human strategic behavior.
\newblock In \emph{Advances in Neural Information Processing Systems}, pages
  2424--2432, 2016.

\bibitem[Hastie and Tibshirani(1990)]{HastieTibshirani:90}
T.~Hastie and R.~Tibshirani.
\newblock \emph{Generalized Additive Models}.
\newblock Chapman Hall, New York, 1990.

\bibitem[Howard and Jebara(2007)]{Jebara:2007}
A.~Howard and T.~Jebara.
\newblock Learning monotonic transformations for classification.
\newblock \emph{Advances in Neural Information Processing Systems {(NIPS)}},
  2007.

\bibitem[Kingma and Ba(2014)]{kingma2014adam}
D.~Kingma and J.~Ba.
\newblock Adam: A method for stochastic optimization.
\newblock \emph{arXiv preprint arXiv:1412.6980}, 2014.

\bibitem[Kondor and Jebara(2003)]{Jebara:03}
R.~Kondor and T.~Jebara.
\newblock A kernel between sets of vectors.
\newblock In \emph{Proceedings of the 20th International Conference on Machine
  Learning (ICML)}, pages 361--368, 2003.

\bibitem[Kotlowski and Slowinski(2009)]{kotlowski2009rule}
W.~Kotlowski and R.~Slowinski.
\newblock Rule learning with monotonicity constraints.
\newblock In \emph{Proceedings of the 26th Annual International Conference on
  Machine Learning}, pages 537--544. ACM, 2009.

\bibitem[Liu et~al.(2015)Liu, Luo, Wang, and Tang]{liu2015faces}
Z.~Liu, P.~Luo, X.~Wang, and X.~Tang.
\newblock Deep learning face attributes in the wild.
\newblock In \emph{Intl. Conf. Computer Vision (ICCV)}, pages 3730--3738, 2015.

\bibitem[Muandet et~al.(2012)Muandet, Fukumizu, Dinuzzo, and
  Schoelkopf]{Muandet:2012}
K.~Muandet, K.~Fukumizu, F.~Dinuzzo, and B.~Schoelkopf.
\newblock Learning from distributions with support measure machines.
\newblock \emph{Advances in Neural Information Processing Systems {(NIPS)}},
  2012.

\bibitem[Poczos et~al.(2012)Poczos, Xiong, Sutherland, and
  Schenider]{Poczos:2012}
B.~Poczos, L.~Xiong, D.~Sutherland, and J.~Schenider.
\newblock Support distribution machines.
\newblock \emph{available on arXiv}, 2012.

\bibitem[Sharma and Bala(2002)]{BalaBook}
G.~Sharma and R.~Bala.
\newblock \emph{Digital Color Imaging Handbook}.
\newblock CRC Press, New York, 2002.

\bibitem[Shivaswamy and Jebara(2006)]{Jebara:06}
P.~K. Shivaswamy and T.~Jebara.
\newblock Permutation invariant svms.
\newblock In \emph{Advances in Neural Information Processing Systems}, 2006.

\bibitem[Sill and {Abu-Mostafa}(1997)]{Sill:97}
J.~Sill and Y.~S. {Abu-Mostafa}.
\newblock Monotonicity hints.
\newblock \emph{Advances in Neural Information Processing Systems {(NIPS)}},
  pages 634--640, 1997.

\bibitem[You et~al.(2017)You, Ding, Canini, Pfeifer, and Gupta]{You:2017}
S.~You, D.~Ding, K.~Canini, J.~Pfeifer, and M.~R. Gupta.
\newblock Deep lattice networks and partial monotonic functions.
\newblock \emph{Advances in Neural Information Processing Systems {(NIPS)}},
  2017.

\bibitem[Zaheer et~al.(2017)Zaheer, Kottur, Ravanbakhsh, Poczos, Salakhutdinov,
  and Smola]{Smola:2017}
M.~Zaheer, S.~Kottur, S.~Ravanbakhsh, B.~Poczos, R.~Salakhutdinov, and
  A.~Smola.
\newblock Deep sets.
\newblock \emph{Advances in Neural Information Processing Systems {(NIPS)}},
  2017.

\end{thebibliography}

\clearpage
\newpage

\section{Appendix A: More Implementation Details}
We provide more details of our implementation, particularly how we constrain the input-and-output range of each layer. 

\subsection{Details on Each Layer}
\textbf{First Layer ($\phi$): Calibrators} Each of the $D$ calibrators has a input range bounded in a user-defined $[\min,\max]$ range, which we set based on the train set or domain knowledge. Each calibrator is a one-dimensional piece-wise linear function stored as a set of $V_d$ pairs of keypoints-and-values, where the keypoints are spaced from $[\min,\max]$ in line with the quantiles of the inputs, the $V_d$ calibrator values get jointly trained but are bounded to $[0,1]$, and the $V_d$ values are initialized to form a linear function spanning the input-output range. Every calibrator is  constrained to be a monotonic function  by adding linear inequality constraints on the adjacent look-up table parameters that restrict each of the one-dimensional look-up table values to be greater than its left neighbor~\citep{GuptaEtAl:2016}. If the $\phi$ function has $K$ outputs, then there are $K \times D$ separate calibrators. 

\textbf{Second Layer ($\phi$): Lattices} The lattice layer follows the calibration layer and thus takes inputs in $[0,1]^{D \times K}$. Each set of $D$ inputs goes into one of $K$ lattices.  The outputs of this lattice layer are all bounded to be in $[-1,1]$, and all the lattices parameters are initialized to $0$. Each lattice is represented by a look-up table with $L_1 \times L_2 \times \ldots L_D$ parameters, where $L_d = 2$ usually suffices, but was tuned on the validation set for some of our experiments to a higher integer value to form a finer-grained lattice. Each lattice is constrained to be monotonic increasing or decreasing with respect to a user-specified set of features by adding appropriate linear inequality constraints on the lattice's look-up table parameters (see \citet{GuptaEtAl:2016} for details). For each experiment, we used a priori domain knowledge to specify the monotonicity constraints. 

\textbf{Third Layer (Simple Average):} The $M(x)$ outputs from each of the $K$ lattices are averaged, such that the average layer outputs $K$ outputs total. We use an average rather than a sum (as in (\ref{eqn:smola})) so that the input(s) to $\rho$ is guaranteed bounded to $[-1,1]$. 

\textbf{Fourth Layer ($\rho$: $K$ Calibrators)}: This layer are largely similar to the first and second layers. A slight difference is that the inputs to the fourth layer lie in $[-1, 1]$, as described above, and that the keypoints are initialized uniformly across the range rather than according to empirical quantiles from the data. Like the other calibrators, their outputs are bounded in $[0,1]$.

\textbf{Fifth Layer ($\rho$: Lattice on $K$ inputs)}: The $K$ calibrated values from the fourth layer get fused together by one $K$-dimensional lattice. We explictly constrain the lattice parameters to all be in $[-1,1]$, and since the output of this layer is just an interpolation of the lattice parameters, that constrains the output of this layer to be in $[-1,1]$.

\textbf{Sixth Layer ($\rho$): Output Calibrator} The last layer is a one-dimensional piece-wise linear transform parameterized by a set of $V_o$ pairs of keypoints-and-values, where the keypoints are uniformly-spaced over its input range $[-1,1]$, and its output range is determined by the training. The $V_o$ values are initialized to be the identity function.  This transform is constrained to be monotonic by adding linear inequality constraints to the training (see (\ref{eqn:erm})) that restrict each of the one-dimensional look-up table values to be greater than its left neighbor~\citep{GuptaEtAl:2016}.

\subsection{Training and Optimization}
Training the proposed aggregation function $f(x; \theta)$ where $\theta$ represents all of the DLN parameters for $\rho$ and $\phi$ is a constrained structural risk minimization problem. Given a training set $\{(x_i, y_i)\}$ for $i = 1, \ldots, n$: 
\begin{equation}
\arg \min_{\theta} \sum_{i=1}^N L \left( f(x_i; \theta), y_i\right)  + R(\theta) \textrm{ such that } G_k(\theta) \leq 0 \textrm{ for all } k = 1, \ldots, K,  \label{eqn:erm}
\end{equation}
where $G$ expresses the monotonicity constraints and any range constraints on the DLN parameters.  

Note the proposed DLN structure for DLN $\rho$ and $\phi$ means all parameters of (\ref{eqn:smola}) have gradients computable with the chain rule and backpropagation.
We solve (\ref{eqn:erm}) using the Light-Touch algorithm~\citep{cotter2016light} to handle the monotonicity constraints on top of Adagrad~\citep{duchi:2011}. Light-Touch samples the constraints, learning which ones are most likely to be violated, and inexpensively penalizes them as the training progresses. It thereby converges to a feasible solution without enforcing feasibility throughout. Once optimization has finished, we project the final iterate, to guarantee monotonicity.

\subsection{Hyperparameter Optimization}
When learning the aggregate function, we tune hyperparameters including learning rate, number of epochs, output dimension $K$ of the calibrated lattice layer $\phi$ that maps $D$ token features into $K$ outputs, and number of keypoints used in each calibration layer in the DLN. We choose the combination of hyperparameters that achieves best model performance measured on the validation dataset.

In the cases when SFE is used to convert sparse categorical features into dense features, we also tune hyperparameters that affect tokenization. Specifically, we tune (a) the maximum size of the created tokens, which means the maximum order of ngrams if the sparse feature is text and the maximum subset size if it is an unordered set of strings; and (b) the filtering criteria of tokens, including the count threshold and the maximum confidence interval width.

\section{Appendix B: Examples}

\begin{figure}[t]
\centering
\begin{tabular}{cc}
\includegraphics[width=0.95\textwidth]{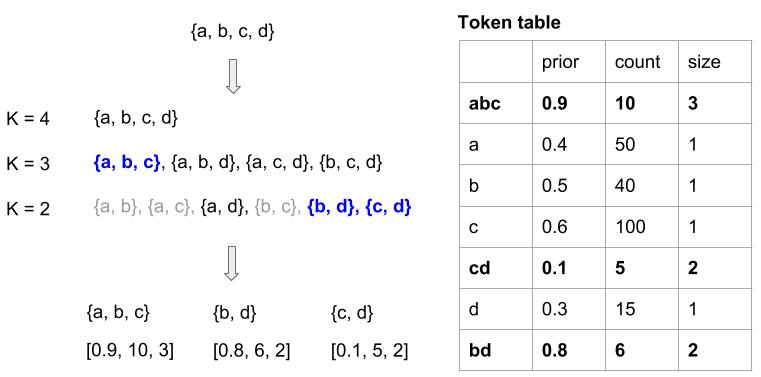}\\
\end{tabular}
\caption{Example SFE tokenization of an input undordered set. It uses a prebuilt token table storing all the filtered tokens and their feature values. The input is a set of 4 categories $\{a, b, c, d\}$. We enumerate subsets of it starting from a maximum subset size $K = 4$, and output ones found in the token table until all categories in the input set are covered. When $K = 4$, we skip the only subset $\{a, b, c, d\}$ because it is not found in the token table. Then we proceed to $K = 3$, out of the 4 subsets we found one $\{a, b, c\}$ in the token table. We add the subset and its features to the output. Since there is one item $d$ in the input set not covered by any subset, we continue with $K = 2$. Out of the 6 subsets of size 2, we skip 3 of them ($\{a, b\}$, $\{a, c\}$, $\{b, c\}$) that are already fully covered by a chosen subset ($\{a, b, c\}$). Out of the rest 3, two of them are found in the token table $\{b, d\}$ and $\{c, d\}$. We output both. Up until now, all items in the input set are covered, so we stop here and output all 3 selected subsets as tokens.}
\label{fig:combinatorial_tokenizer} 
\end{figure}

\subsection{Example of SFE Fallback Logic}
Fig. \ref{fig:combinatorial_tokenizer} illustrates how the SFE fallback logic works when we generate a set of dense feature vectors from an un-ordered set of categories. The general idea is to find large subsets of the input to cover all the categories, and fall back to their smaller subsets if they are not found in a pre-built token table. A subset does not make to the token table if it does not appear frequent enough in the training data according to the filtering criteria.

\subsection{Complete SFE and Learned Aggregation Function Example}
We show the inner workings of the aggregation function on an example from the recipes dataset. The example, whose true cuisine is French but for which we are evaluating the candidate cuisine of Mexican, consists of the seven ingredients \{sugar, salt, fennel bulb, water, lemon olive oil, grapefruit juice\}. As mentioned above, we consider subsets of up to size 3 for this problem. As the combination of ingredients is rather rare, we only find 2 subsets of size 3 (out of $7 \choose 3$ possibilities) that appeared in the training data at least 5 times and are therefore in the token table: \{salt, water, fennelbulb\} and \{salt, water, sugar\}. Since neither lemon olive oil nor grapefruit juice appear in any frequent subsets of size 3, the model searches for any subsets of size 2 containing one of those ingredients and doesn't find any; finally it finds grapefruit juice as a singleton in the token table, while lemon olive oil never appears in any form in the token table.

We therefore have the following 3 total tokens:
\begin{enumerate}
\item \{salt, water, fennel bulb\}: $P\left\{\textrm{Mexican}\right\} = 0.059$; count = 17; subset size = 3; is full match = 0; number of ingredients: 6; number of tokens: 3
\item \{salt, water, sugar\}: $P\left\{\textrm{Mexican}\right\} = 0.043$; count = 510; subset size = 3; is full match = 0; number of ingredients: 6; number of tokens: 3
\item \{grapefruit juice\}: $P\left\{\textrm{Mexican}\right\} = 0.2$; count = 10; subset size = 1; is full match = 0; number of ingredients: 6; number of tokens: 3
\end{enumerate}
Recall that there are 20 possible cuisines, so 0.05 is the break-even uninformative prior value. Here, both subsets of size 3 have fairly weak/neutral evidence while the prior for grapefruit juice is actually quite high.

The intermediate outputs $\phi(x)$ for these tokens are:
\begin{enumerate}
\item \{salt, water, fennel bulb\}: -0.005
\item \{salt, water, sugar\}: -0.156
\item \{grapefruit juice\}: 0.075
\end{enumerate}
Our labels are -1/1, so positive outputs represent the model leaning towards yes and negative outputs are the opposite. Interestingly, the \{salt, water, sugar\} subset is much more negative than \{grapefruit juice\} is positive, even though its prior is much closer to neutral. We can explain this by looking at the supporting information: the former has a subset size of 3, teaching the model to trust it more than a subset of size 1, and it appears 510 times vs. 10, leading to the same conclusion. 

Finally, the outputs are averaged together and fed through $\rho$, which in the $K=1$ case is a simple piecewise linear transform, yielding the final output of -0.28. The classifier has correctly identified the recipe as not being Mexican.

\end{document}